\documentclass[runningheads]{llncs}
\usepackage{graphicx}

\usepackage{cite}
\usepackage{amsmath}
\usepackage{siunitx}
\usepackage{glossaries}
\newacronym{model}{MME}{Multi-magnification Ensemble}
\newacronym{ensemble}{SME}{Single-magnification Ensemble}
\newacronym[longplural={Vision Transformers}, shortplural={ViTs}]{vit}{ViT}{Vision Transformer}
\newacronym[longplural={Convolutional Neural Networks}, shortplural={CNNs}]{cnn}{CNN}{Convolutional Neural Networks}
\newacronym[longplural={Whole Slide Images}, shortplural={WSIs}]{wsi}{WSI}{Whole Slide Image}
\newacronym{mil}{MIL}{Multiple Instance Learning}
\newacronym{hae}{H\&E}{Hemaoxylin and Eosin}
\newacronym{mcc}{MCC}{Matthews Correlation Coefficient}
\newacronym{auroc}{AUROC}{Area Under the Receiver Operating Characteristic}
\newacronym{nlp}{NLP}{natural language processing}
\newacronym{mme}{MME}{Multi-magnification Ensemble}
\newacronym[first={CAMELYON16}]{dataset}{CAMELYON16}{CAMELYON16}
\usepackage{bm}

\usepackage[pagebackref,breaklinks,colorlinks]{hyperref}
\usepackage{cleveref}
\usepackage{soul}
\usepackage{caption}
\usepackage{subcaption}
\newlength{\twosubht}
\newsavebox{\twosubbox}

\begin{document}
\title{Robust Tumor Detection from Coarse Annotations via Multi-Magnification Ensembles}
\titlerunning{Robust Tumor Detection from Coarse Annotations}
\author{Mehdi Naouar\inst{1,2} \and
Gabriel Kalweit\inst{1,2} \and Ignacio Mastroleo\inst{2,3,4} \and Philipp Poxleitner\inst{2,5} \and Marc Metzger\inst{6} \and
Joschka Boedecker \inst{1,2} \and Maria Kalweit\inst{1,2}}
\institute{University of Freiburg \and Collaborative Research Institute Intelligent Oncology (CRIION) \and Program of Bioethics, Facultad Latinoamericana de Ciencias Sociales (FLACSO) \and 
National Scientific and Technical Research Council (CONICET)\and University Hospital of Munich (LMU) \and Medical Center Freiburg}
\maketitle              % typeset the header of the contribution
\begin{abstract}
Cancer detection and classification from gigapixel whole slide images of stained tissue specimens has recently experienced enormous progress in computational histopathology. The limitation of available pixel-wise annotated scans shifted the focus from tumor localization to global slide-level classification on the basis of (weakly-supervised) multiple-instance learning despite the clinical importance of local cancer detection. However, the worse performance of these techniques in comparison to fully supervised methods has limited their usage until now for diagnostic interventions in domains of life-threatening diseases such as cancer. In this work, we put the focus back on tumor localization in form of a patch-level classification task and take up the setting of so-called coarse annotations, which provide greater training supervision while remaining feasible from a clinical standpoint. To this end, we present a novel ensemble method that not only significantly improves the detection accuracy of metastasis on the open CAMELYON16 data set of sentinel lymph nodes of breast cancer patients, but also considerably increases its robustness against noise while training on coarse annotations. Our experiments show that better results can be achieved with our technique making it clinically feasible to use for cancer diagnosis and opening a new avenue for translational and clinical research.

\end{abstract}
\section{Introduction}
\label{sec:intro}
\begin{figure}[t]
    \centering
    \includegraphics[width=\textwidth]{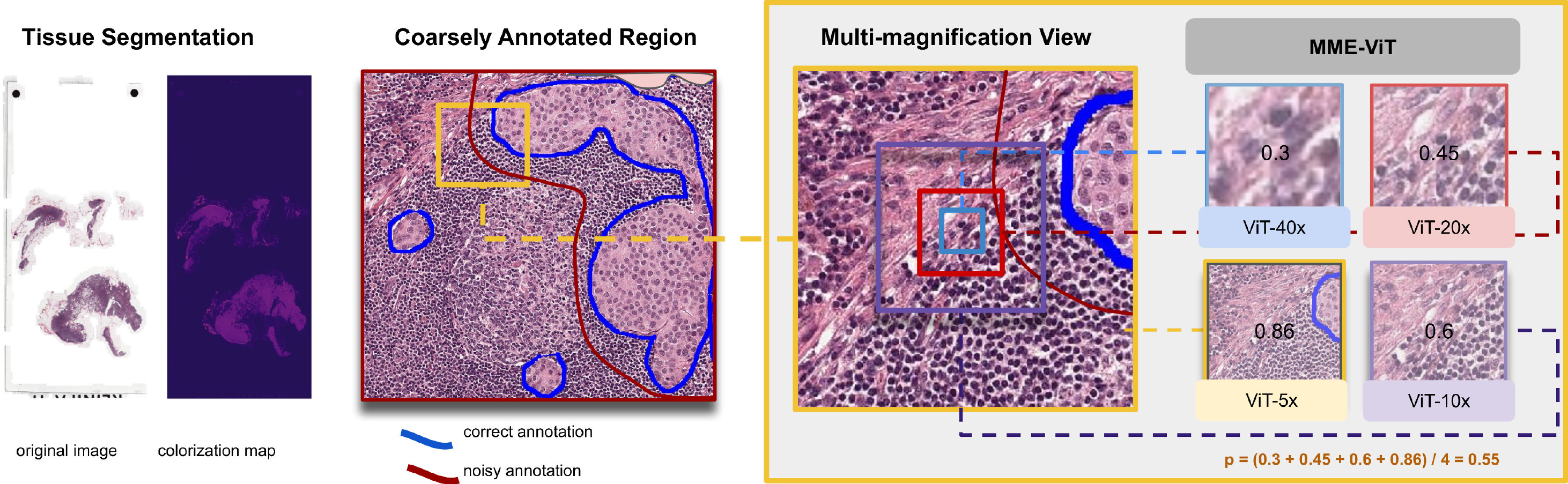}
    \caption{Robust tumor detection by averaging over predictions from \glsentrylongpl{vit} trained on multiple magnifications to account for coarse annotations.
    }
    \label{fig:coverlady}
\end{figure}

In recent years, computational histopathology achieved tremendous successes in the support of medical diagnosis and prognosis from gigapixel \glspl{wsi}~\cite{Chen_2022_CVPR,Lu2021,Liu2017,Li_2021_CVPR}. Despite the unique challenges arising from the sheer size of such scans, typically in the range of $10^5\times10^5$ pixels, approaches on the basis of deep neural networks gained more and more traction and made the employment of sophisticated non-linear function approximation feasible. One very promising application of such techniques is the detection and classification of cancer in scans of \gls{hae} stained tissue specimens. Since pixel-wise annotations require prohibitively many man hours to obtain, the research focus shifted from tumor localization to global slide-level classification on the basis of (weakly supervised) \gls{mil} techniques. Even though such approaches led to decent results on \textit{global localization} of cancerous tissue~\cite{Chen_2022_CVPR},  better supervision in form of (pixel-wise) annotations can still improve the classification and localization of tumorous regions~\cite{pmlr-v156-tourniaire21a,https://doi.org/10.48550/arxiv.2012.03583}. We hence propose to instead leverage \textit{coarse} annotations which offer better supervision compared to slide-level labels for the localization of tumor cells while remaining feasible from a clinical perspective, typically reducing the time required from 5 to 6 hours \cite{LINDMAN201922, SCHUFFLER20219} to 10 minutes per \gls{wsi}. The assumption of coarse annotations, however, leads to noisy ground truths engendering a subset of miss-labeled samples while training a patch classifier. Our goal is thus to develop a model robust to the noise arising from the nature of these annotations while still being capable of exploiting their concomitant benefits. In this perspective, different magnifications may imply different but \emph{complementary} features, mostly due to their respective unique fields-of-view which yields the possibility to furthermore enhance the robustness and lead to better class separation in both, noisy and non-noisy settings. On that account, we hypothesize that multi-magnification ensembles bear the potential to surpass the performance of their single-magnification counterparts in this very domain.
Our contributions are threefold. First, we define an analysis and a consistent, non-biased evaluation for local tumor detection at multiple magnifications and show that models trained on different magnifications have different strengths and weaknesses on different scales. Second, we present a novel \gls{model} for patch-level tumor classification (cf. \Cref{fig:coverlady}). Third, we evaluate our approach on the open \gls{dataset} dataset, showing that taking into account information at multiple magnifications can lead to better localization results and to more robustness in the presence of training noise.

\section{Related Work}
\label{sec:relatedwork}
\textbf{Model-ensembles.}
There is a great body of work on combining several individual models in an ensemble to obtain better generalization performance~\cite{googlenet,LeePCCB15, ensemblelosslandscape}. Deep ensemble-learning methods have recently entered the medical domain~\cite{Cao2020}, e.g. to predict the SPOP mutation state in prostate cancer via an ensemble of ResNets~\cite{Schaumberg064279}, for active learning on histopathology images of breast cancer samples \cite{lee2019ensemble} or to identify APL in bone marrow smears \cite{Eckardt2022}.

\textbf{Coarse Annotations.}
Several label denoising methods~\cite{labelcleaning, Ashraf2022} have been proposed with the same motivation of avoiding costly pixel-wise annotations by cleaning coarse annotations. While \cite{labelcleaning} applies label denoising using multiple instance learning, \cite{Ashraf2022} detects mislabeled patches based on the training loss of the single patches during training. While these methods have the same motivation of enabling the training on coarse annotations, they are orthogonal to our work and could be combined with our approach to reach a better performance.

\textbf{Multi-magnification Methods.}
 Multi-magnification \glsentrylongpl{cnn} were used for multi-class breast cancer image segmentation in~\cite{Liu2017,Ho_2021, Alfonso2021}, exploiting the lower magnification to have wider field-of-view by concatenating the vector embeddings of lower magnifications. In~\cite{Chen_2022_CVPR}, a hierarchical stacking of embeddings is used to aggregate  visual tokens at different magnifications to form slide representations. These methods fuse information from multiple magnifications in the input space and follow a similar rationale as our work. In contrast to our proposed approach, however, they incorporate this information to end up with only one \emph{single} estimate. As has been shown in prior work, ensemble methods not only lead to better performance overall compared to their single-model counterparts~\cite{googlenet,LeePCCB15}, they also offer better uncertainty-estimates and enhanced robustness in terms of better calibration~\cite{ensembleuncertainty}. We therefore argue that ensemble methods are a better fit in the presence of noise of coarse annotations which we consider a more feasible setting from a real-life clinical perspective. %Thus, this particular set of attributes, (1) ensembles over (2) multiple magnifications to allow for (3) coarse annotations.

\section{Method}
\label{sec:method}

In the following, we explain the single steps of our method. We present a novel way of tissue segmentation, formalize the task of supervised patch-level classification, and provide a detailed description of our proposed \glsentrylongpl{model}.

\subsection{Tissue Segmentation}
 To reduce computation time and increase the data efficiency during training, we first identify tissue within the \gls{wsi} and extract patches exclusively within this region of interest.
While most works rely on a tissue segmentation method using the adaptive thresholding over the saturation channel in the HSV color space~\cite{https://doi.org/10.48550/arxiv.1606.05718,Khened2021,Lu2021},  this method may fail while filtering out the dark background noise. To address this issue, we present a novel unsupervised tissue segmentation approach. % Since in some of the \gls{dataset} cases, the  thresholding~\cite{https://doi.org/10.48550/arxiv.1606.05718,Khened2021,Lu2021}  in HSV color-space failed because of the black regions in the \gls{wsi},
After computing an image mask using the \emph{colorization value} defined as $c(r, g, b) = |r-m| + |g-m| + |b-m|,$ where $m = \frac{r+g+b}{3}$, we apply an adaptive threshold~\cite{Otsu1979ATS} to separate the tissue from the background. The colorization value of a pixel is lowest for all shades of grey and is high for all others.
Our technique recognizes tissue regions accurately and succeeds in segmenting out some parts of the background, whereas the saturation-based thresholding used to fail (see \cref{fig:segmentation}).

\subsection{Supervised Patch-level Classification}
Then, patches $p^{\mu_k, s_k}_{\omega,i}$ of magnification $\mu_k \in \{40\times, 20\times, 10\times, 5\times\}$ with corresponding patch size  $s_k \times s_k$  (i.e $s_k \in \{256, 512, 1024, 2048\}$) before down-sampling are extracted from the segmented tissue region of each \gls{wsi} $\omega$. We denote the ground truth label of a patch $p^{\mu_k, s_k}_{\omega,i}$ by $y^{s_k}_{\omega,i} \in \{0, 1\}$ indicating weather it contains tumor cells. The total set of labeled patches $D^{\mu_\text{k}} = \{\bigcup (p_{\omega,i}^{\mu_\text{k}}, y_{\omega,i}^{s_\text{k}}) | \forall \omega,i \}$  yields an imbalanced distribution since tumorous tissue typically represents a small portion of the tissue region in a \gls{wsi} inducing a strong class imbalance between patches of the positive and negative class.
We address this issue by under-sampling the majority class.  A heavy color augmentation is applied to the input patches during training to remedy the stain variations of different \glspl{wsi}.

\textbf{Learning with Noisy Labels.} To enhance clinical feasibility, we lift the assumption of pixel-wise annotations by artificially generating  coarse annotation through two operations. First, the tumor polygons of the pathologists are expanded, increasing the number of false positive labeled patches around the real tumor regions. Second, some tumor annotations are randomly discarded with probability $\eta$ producing false negative samples in the training set. 
\subsection{\glsentrylongpl{mme}}
\begin{figure*}[b]
    \centering
    \includegraphics[width=1\textwidth]{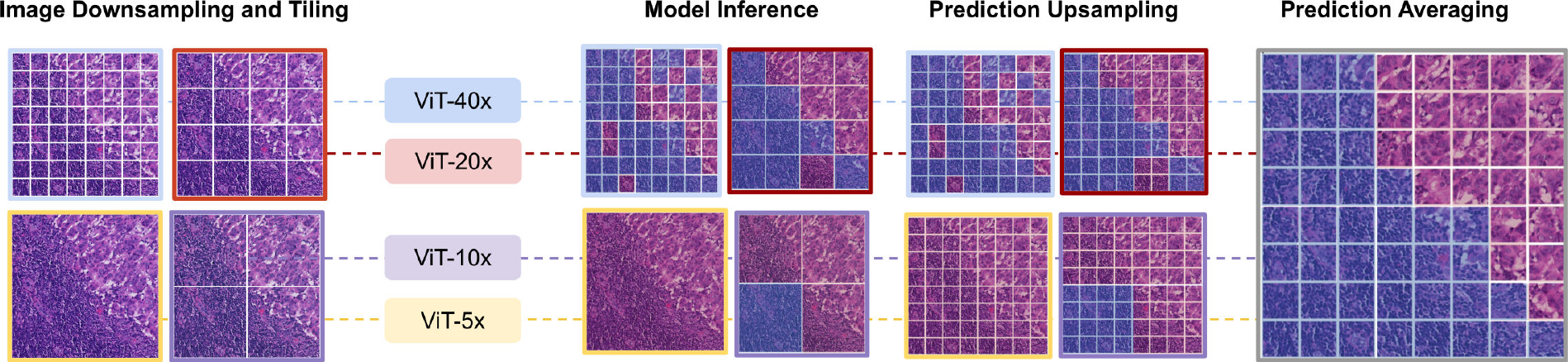}
    \caption{Inference pipeline of the \glsentrylong{model}.
    }
    \label{fig:ensemble}
\end{figure*}
To exploit the robustness of ensembles and to account for label-noise, we suggest using multi-magnification ensembles of \glspl{vit} on a set of magnifications $(\mu_1, ..., \mu_K )$ with $\mu_1 <  \cdots < \mu_K$. Our ensemble method is motivated by the trade-off set by the magnification between the resolution and the field of view of the classifier. On the one hand, using high magnification enables a better distinction of small structures. On the other hand, using a lower magnification enlarges the field of view of the network at the cost of image resolution. We denote a \gls{vit} trained on inputs patches of magnification $\mu_k$ by $\text{\gls{vit}-}\mu_k$, and its prediction on a patch $p_{\omega,i}^{\mu_k, s_k}$ by $\hat{y}_{\omega,i}^{\mu_k, s_k}$. During the inference, each model provides a prediction over all patches within the tissue region. Although the patches of different models cover the exact same area, these differ in size $s_k \times s_k$. Therefore, the single prediction areas of different models have different sizes. We thus rescale the predictions using up-sampling to match the smallest patch size $s$, i.e. $\hat{y}_{\omega,i}^{\mu_k, s_k} \mapsto \hat{y}_{\omega,i}^{\mu_k, s}$, as illustrated in \Cref{fig:ensemble}. At last, the up-sampled predictions are combined using the uniform average: $\hat{y}_{\omega,i}^{e,s} =  \frac{1}{K} \sum_{k=1}^K \hat{y}_{\omega,i}^{\mu_k, s}$.

\section{Experiments}
\label{sec:experiments}

\textbf{Experimental Setup.} We evaluated the patch classification performance of our models on the open \gls{dataset}  dataset~\cite{Litjens20181399HS}, containing 399 \gls{hae} stained \glspl{wsi} of sentinel lymph node sections from breast cancer patients together with annotations by expert pathologists. The training set contains 159 normal (tumor-free) slides and 111 tumor slides (containing macro- and/or micro-metastasis). The independent test set contains 81 normal and 48 tumor slides. We make use of \gls{mcc} in our evaluation as it has been shown to be a more suited performance measure for binary classification tasks in the presence of class imbalance. In contradiction to the F1-Score, \gls{mcc} takes into account all entries of the confusion matrix making it invariant to the class distribution~\cite{Chicco2020}. Moreover, we report the \gls{auroc} performance to measure the separability of the positive and negative classes over different classification thresholds.
Besides undersampling the majority class, we apply data augmentation by uniformly sampling transformation on the brightness, contrast, saturation, and hue of the input patches.  Furthermore, we apply random flipping, as the orientation of a tissue sample should not be relevant for the detection of tumor cells. All models have been pre-trained on ImageNet~\cite{deng2009imagenet} and trained on patches of \glspl{wsi} from \glsentryshort{dataset} for 5000 iterations with a batch size of 256 using an Adam optimizer \cite{adam} with parameters $\beta$=(0.9, 0.999) and $\epsilon=10^{-8}$. We make use of a cosine annealing learning rate scheduler with a linear learning rate warm-up over the first $\frac{1}{6}$ of the total training iterations, a maximal learning rate of $10^{-4}$ and a minimal learning rate of $10^{-5}$.\\

\begin{figure}[htp]

\sbox\twosubbox{%
  \resizebox{\dimexpr.99\textwidth-1em}{!}{%
    \includegraphics[height=2.3cm]{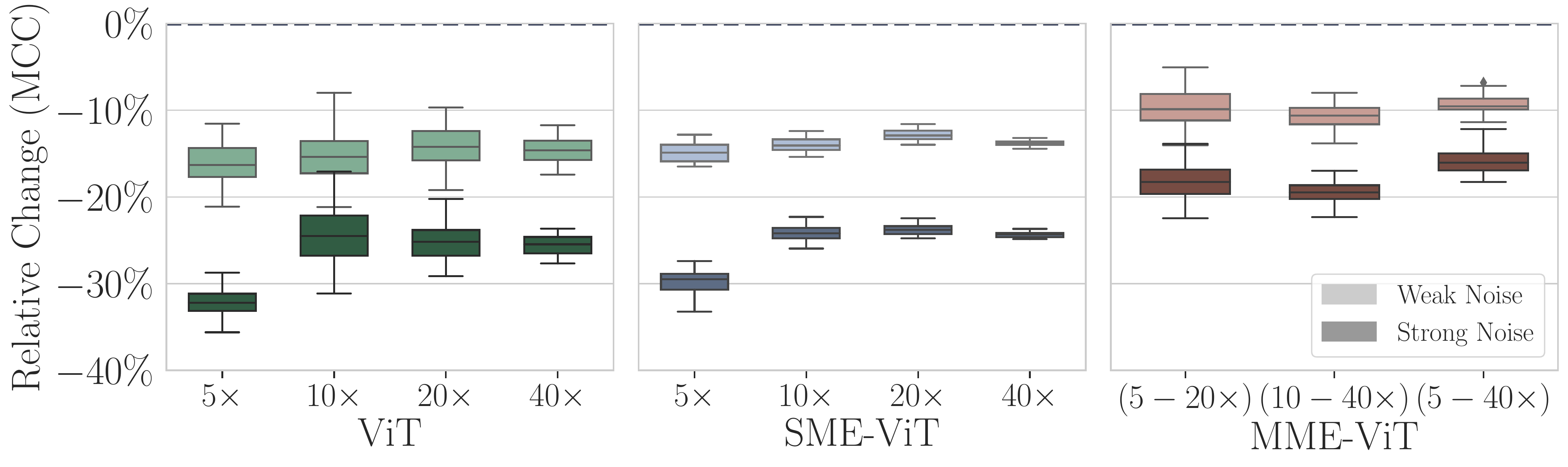}%
    \includegraphics[height=2.3cm]{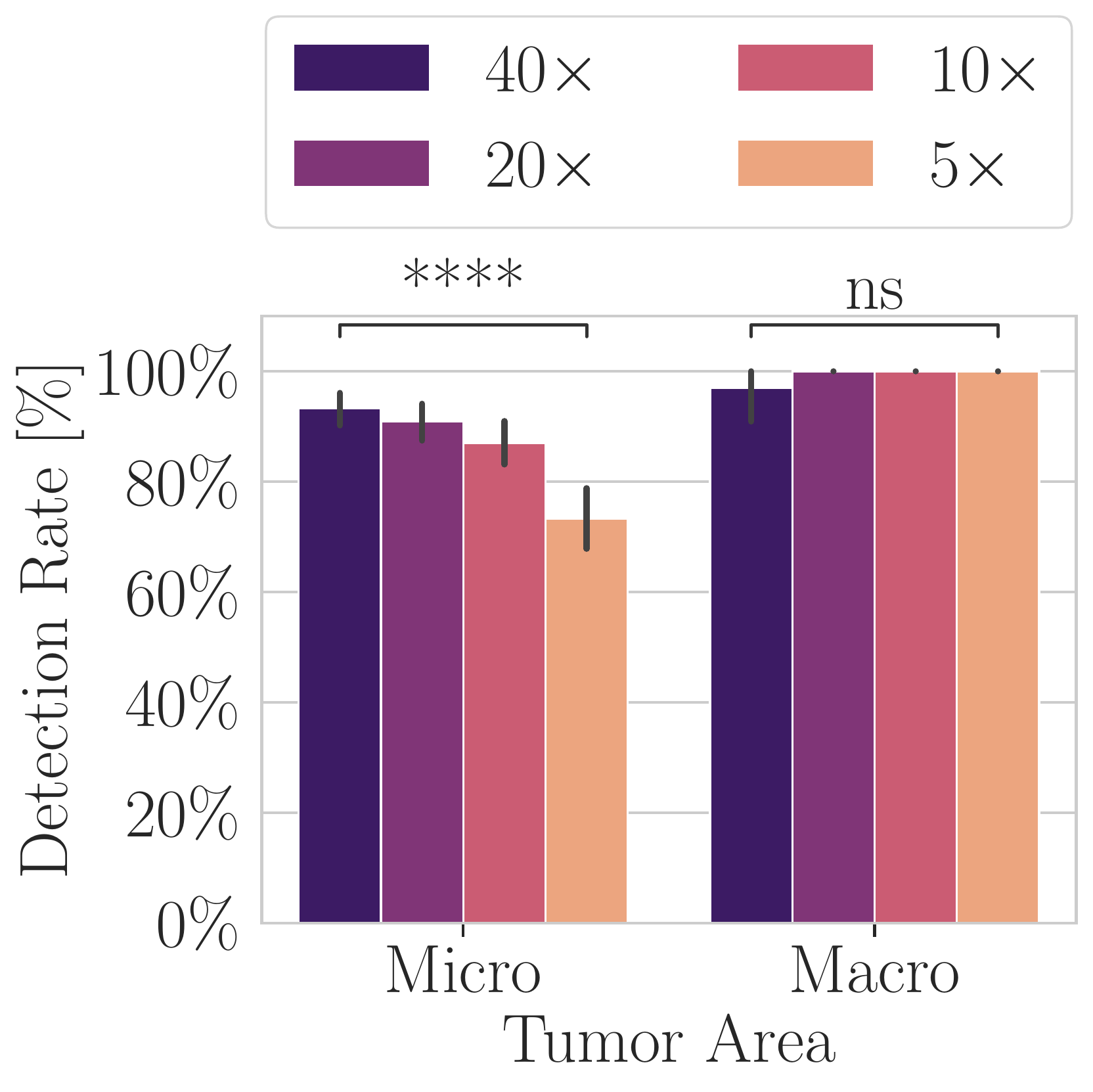}%
  }%
}
\setlength{\twosubht}{\ht\twosubbox}

\centering

\subcaptionbox{\label{f}}{%
  \includegraphics[height=\twosubht]{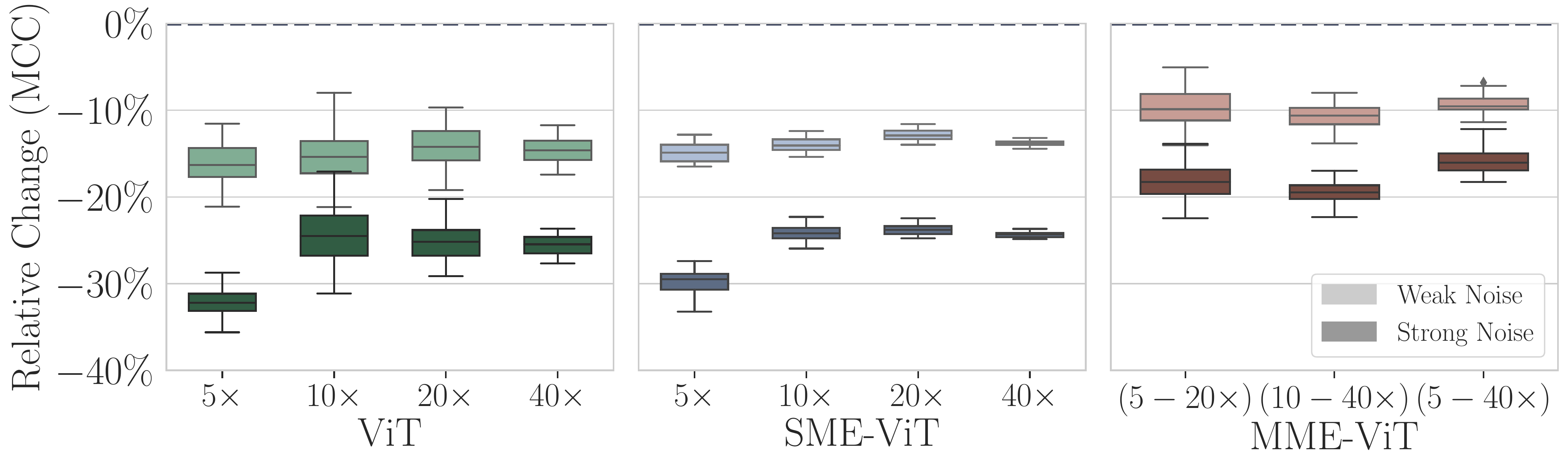}%
}\quad
\subcaptionbox{\label{s}}{%
  \includegraphics[height=\twosubht]{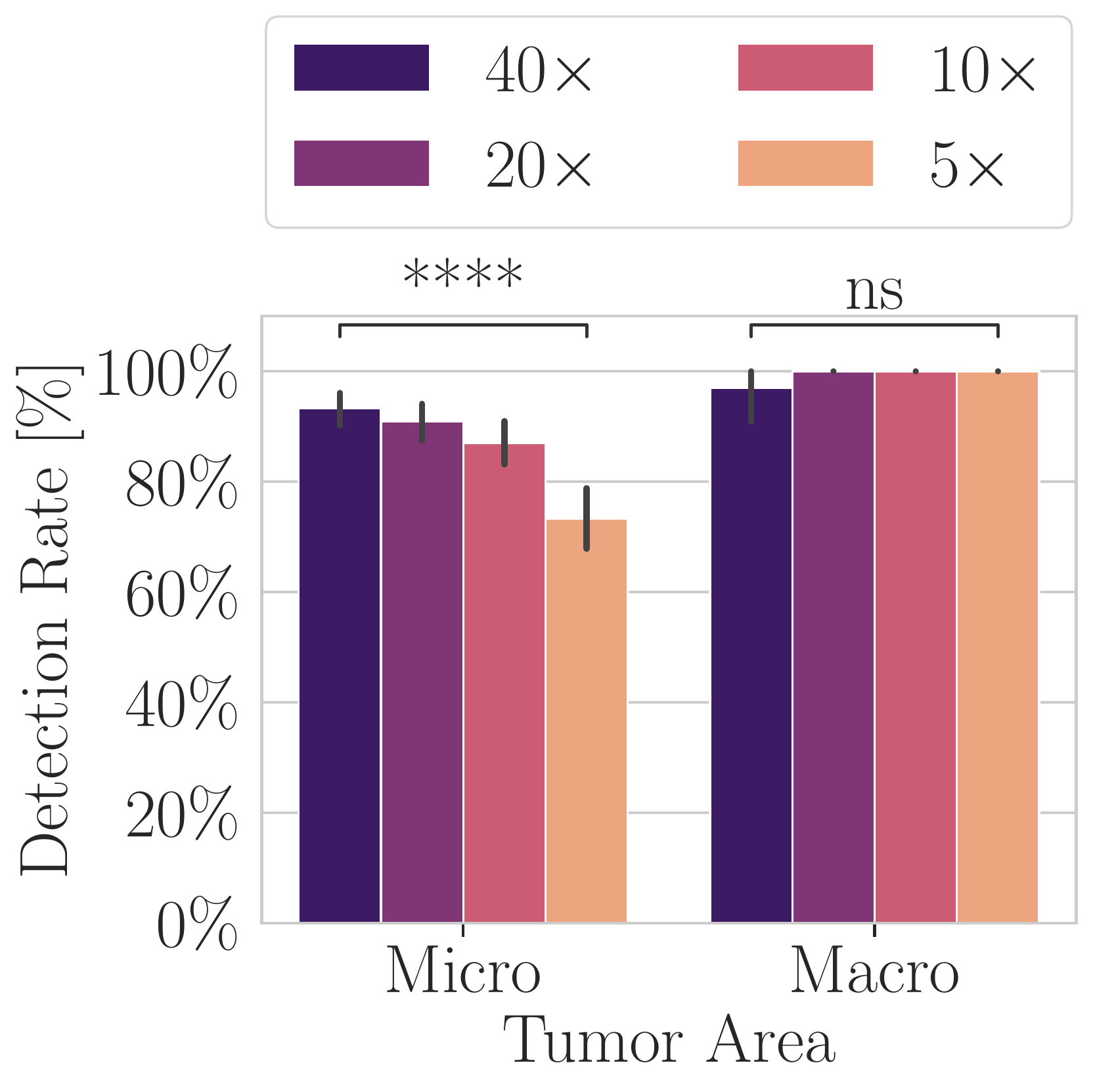}%
}

\caption{(a) Relative change in \gls{mcc} when training with \emph{weak noise} and \emph{strong noise} for \glspl{model} and all baselines. (b) Tumor detection rate of micro- and macro-metastasis for the models trained on input magnifications $40\times, 20\times, 10\times$ and $5\times$.}
\label{fig:noise_rel_aligned_MCC}

\end{figure}

\textbf{Results.}
To test our initial hypothesis and motivate the benefit of a multi-scale ensemble, we evaluate the detection performance of different magnification models on (small) micro- and (large) macro-metastasis. We trained \glspl{vit} on different input magnifications $\mu_k \in \{5,10,20, 40\}$ and corresponding patch sizes of $s_k\in\{2028,1024,512, 256\}$ and measured their detection rate on micro- and macro-metastasis, which is defined as the portion of tumors detected by the classifier. We show the tumor detection rate in \Cref{s}. We observed that the detection rate for micro-metastasis improves the higher the magnification, which motivates the need for \glspl{vit} at smaller scales. However, further investigations showed that despite all macro-metastasis being recognized by all models, the portion of detected larger macro-metastasis increased the lower the magnification which explains the lower recall of the \gls{vit}-$40\times$ model in \Cref{tab:evalnonoise}.\\

Next, we combine the different magnification models into three types of ensembles: $( 5\times, 10\times, 20\times)$, $(10\times, 20\times, 40\times)$ and $(5\times, 10\times, 20\times, 40\times)$ and compare them to single \glspl{vit}-$\mu_k$ models, and to \glspl{ensemble}-$\mu_k$  consisting of $l$ \glspl{vit} trained on the corresponding magnification $\mu_k$. To ensure a fair comparison over different tumor sizes, we compute the average performance over all slides to avoid the results being dominated by a subgroup of the slides (i.e. slides having very large or very small or no tumorous regions). An evaluation taken over five different seeds can be seen in \Cref{tab:evalnonoise}. The best \gls{model} with $(5\times, 10 \times, 20\times, 40\times)$ is outperforming the single-magnification models \glspl{vit} and the single-magnification model-ensembles \glspl{ensemble} highly significantly with  p-values $p \leq 10^{-4}$ determined by Welch's t-test, as can also be seen in \cref{fig:significance}.

\begin{table*}[t]
		\centering
		\begin{tabular}{c|ccc|cc}
			\hline
	Configuration    & Precision & Recall & Specificity & \textbf{\gls{auroc}} &\textbf{\gls{mcc}}\\
            \hline
\glsentryshort{vit}-$5\times$ & $34.68 \pm 0.77$ & $71.97 \pm 1.3$ & $97.59 \pm 0.13$ & $92.79 \pm 0.29$ & $42.74 \pm 0.83$\\
\glsentryshort{vit}-$10\times$ & $41.09 \pm 2.28$ & $79.12 \pm 0.99$ & $98.16 \pm 0.23$ & $95.67 \pm 0.17$ & $50.25 \pm 2.02$\\
\glsentryshort{vit}-$20\times$ & $48.02 \pm 1.73$ & $80.23 \pm 0.81$ & $98.72 \pm 0.14$ & $96.3 \pm 0.1$ & $55.16 \pm 1.11$\\
\glsentryshort{vit}-$40\times$ & $50.77 \pm 0.98$ & $68.48 \pm 0.97$ & $99.09 \pm 0.1$ & $93.83 \pm 0.26$ & $52.39 \pm 0.7$\\

	\hline
    \glsentryshort{ensemble} ($5\times^4$) & $36.66 \pm 0.31$ & $72.89 \pm 0.75$ & $97.89 \pm 0.04$ & $93.72 \pm 0.1$ & $45.1 \pm 0.5$\\
\glsentryshort{ensemble} ($10\times^4$) & $43.64 \pm 0.66$ & $79.68 \pm 0.48$ & $98.47 \pm 0.06$ & $96.1 \pm 0.06$ & $52.59 \pm 0.5$\\
\glsentryshort{ensemble} ($20\times^4$) & $51.26 \pm 0.52$ & $80.5 \pm 0.27$ & $98.97 \pm 0.03$ & $96.62 \pm 0.01$ & $57.71 \pm 0.31$\\
\glsentryshort{ensemble} ($40\times^4$) & $53.6 \pm 0.26$ & $68.45 \pm 0.38$ & $99.25 \pm 0.02$ & $94.23 \pm 0.06$ & $54.39 \pm 0.18$\\
    \hline
    \glsentryshort{model} ($40, 20, 10$)  & $62.66 \pm 1.28$ & $79.01 \pm 0.58$ & $99.54 \pm 0.04$ & $97.76 \pm 0.06$ & $65.8 \pm 0.99$\\
 \glsentryshort{model} ($20, 10, 5$)& $54.82 \pm 1.57$ & $79.17 \pm 0.74$ & $99.13 \pm 0.05$ & $97.73 \pm 0.09$ & $60.99 \pm 1.18$\\
 \gls{model} ($40, 20, 10, 5$)& $64.15 \pm 1.22$ & $78.81 \pm 0.61$ & $99.44 \pm 0.01$ & $\bm{98.15 \pm 0.04}$ & $\bm{67.27 \pm 0.64}$\\
                        \hline

		\end{tabular}
				\caption{Classification performance averaged over all \glspl{wsi} over five runs on the \gls{dataset} test set for \glspl{model} and all baselines. $^4$ denotes single-magnification ensembles over 4 models acting on the same scale.}
        \label{tab:evalnonoise}
\end{table*}

Furthermore, we introduce two different noise levels --  \emph{weak} noise and \emph{strong} noise -- for the artificially simulated coarse annotations and train all models on the resulting noisy datasets. The evaluation is performed on the test set without any noise to compare against the ground truth.  The relative performance drop of all models induced by the two noise levels is shown in \Cref{f}. All \glspl{model} have a significantly lower drop in performance compared to the single-magnification \glspl{vit} and their respective \glsentrylongpl{ensemble}.
The results of the models trained on coarse annotations are shown in \cref{tab:evalnoise}. \gls{model} outperforms all other approaches in both \gls{auroc} and \gls{mcc} with significance levels of $p\leq10^{-3}$ ($p \leq 10^{-4}$ in the majority of evaluations, determined by Welch's t-test).%, as shown in \Cref{fig:mccaurocwsi} (c) and (d).

\section{Discussion}
\label{sec:discussion}

\begin{figure}[t]
    \centering
    \includegraphics[width=1\textwidth]{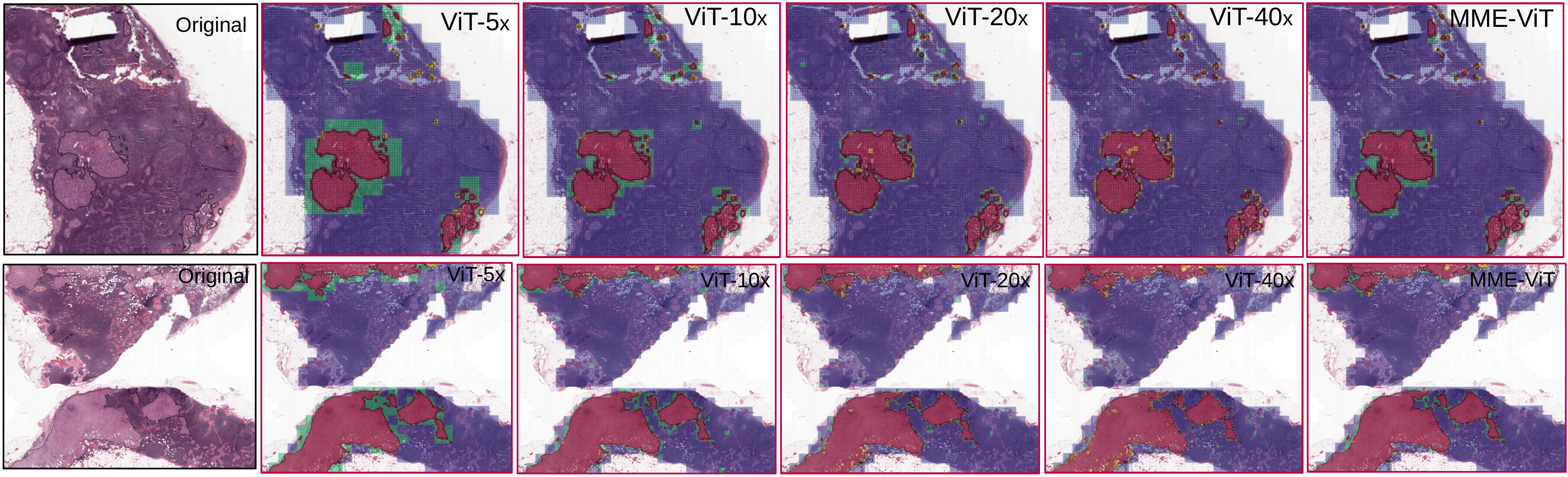}
    \caption{Exemplary evaluation of all \gls{vit}-$\mu_k$ and the resulting \gls{model}. (red) True positives, (blue) true negatives, (yellow) false negatives, (green) false positives.}
    \label{fig:sample}
\end{figure}

From the results in \Cref{s}, it appears that smaller tumors are harder to detect from afar, yet larger ones are easier to detect with a larger field of view at low magnifications. Micro-metastasis are, however, of utmost clinical relevance as these are challenging to detect by pathologists, where computational methods could contribute the most. Consequently, it is beneficial to rather take into account all magnifications to reach a conclusion (cf. \Cref{fig:sample}). This is in line with our findings in \Cref{tab:evalnonoise}, where \glspl{model} improve upon single-scale ensembles by at least $3\%$ in terms of \gls{mcc} and \gls{auroc}. An even larger improvement can be seen when all different magnifications are combined in the same ensemble. It has to be pointed out, however, that, in contrast to prior work \cite{https://doi.org/10.48550/arxiv.1606.05718,Liu2017,Lu2021}, we find the highest magnification ($40\times$) to underperform lower magnifications in many cases. Interestingly, this result implies that a computational tool for diagnosis does not necessarily profit from the highly-detailed features offered at high magnifications alone. Our experimental conclusion is thus consistent with how human pathologists operate, who seldom zoom in to the highest resolution setting \cite{ASHMAN2022100113}.\\

\textbf{Limitations.}
The downside of combining the complementary strengths of models at different scales is that this goes hand in hand with combining their complementary weaknesses which translates to decreasing performance on some \glspl{wsi} in some cases. However, to this end, the benefits outweigh the detriments.
One way to approach this issue would be to identify failure cases and leverage a more sophisticated combination rather than plain averaging. Having such failure modes is especially harmful if medical professionals take predictions and suggestions of a model as being true without question. We want to emphasize that our model, like most other diagnosis assistant systems resulting from current research, should only be seen as complementary support to a human domain expert making the final prediction.

\section{Broader Impact}
Cancer is the second leading death cause worldwide~\cite{ritchie_causes_2019}. In 2020, there were 10 million cancer-related deaths, increasing to 16.3 million expected deaths by 2040~\cite{sung_global_2021}. Cancer is also known as \emph{the emperor of all maladies}~\cite{mukherjee_emperor_2011} for its inherent biological complexity, making it a suitable domain for applying learning systems that can help reduce this global public health burden. 
In this paper, we have developed a proof of concept of a novel algorithm (\gls{model}) that enhances stained cancer tissue detection over standard techniques with the potential of being clinically useful, cost-effective, and scalable. Our research adds a novel tool that could be applied to different problems of stained pathological tissue classification. Moreover, \gls{model} may open a pathway for translational and clinical research of diagnostic devices for cancer and other serious or life-threatening diseases. 

\section{Conclusion}
\label{sec:conclusion}
\glsresetall
In this study, we analyzed the influence of magnification in patch-level classification from gigapixel \glsentryshortpl{wsi} of \glsentryshort{hae} stained tissue specimen. In our experimental evaluation on the \glsentryshort{dataset} dataset, we have seen qualitative differences regarding tumors of different sizes. Surprisingly, our findings shed light on the fact that especially the widely-used $40\times$-magnification underperforms its lower-magnification companions in many use cases, except for very small metastases. This motivated the development of \gls{model}. We have shown that combining different predictions in an ensemble over multiple magnifications and therefore unifying their complementary strengths can increase prediction performance and enhance the robustness against noise significantly. Our analysis of different magnification-combinations within \gls{model} has shown that the additional features extracted at high magnification \emph{can} help but have the least impact on average which is in line with recent results on the workflow of human pathologists \cite{ASHMAN2022100113}. Going forward, extending the analysis to other forms of malignancies could further demonstrate the underlying potential of our proposed method, especially since we were aiming at clinical feasibility with our initial motivation of coarse annotations. %Secondly, we believe that other ways of pooling beyond averaging could be a very promising avenue, e.g. by combining multiple \glsentrylongpl{ensemble} on different magnifications and estimating their respective uncertainties. 

\bibliographystyle{splncs04}

\begin{thebibliography}{10}
\providecommand{\url}[1]{\texttt{#1}}
\providecommand{\urlprefix}{URL }
\providecommand{\doi}[1]{https://doi.org/#1}

\bibitem{ASHMAN2022100113}
Ashman, K., Zhuge, H., Shanley, E., Fox, S., Halat, S., Sholl, A., Summa, B.,
  Brown, J.Q.: Whole slide image data utilization informed by digital diagnosis
  patterns. Journal of Pathology Informatics  \textbf{13},  100113 (2022)

\bibitem{Ashraf2022}
Ashraf, M., Robles, W.R.Q., Kim, M., Ko, Y.S., Yi, M.Y.: A loss-based patch
  label denoising method for improving whole-slide image analysis using a
  convolutional neural network. Scientific Reports  \textbf{12}(1), ~1392 (Jan
  2022)

\bibitem{Cao2020}
Cao, Y., Geddes, T.A., Yang, J.Y.H., Yang, P.: Ensemble deep learning in
  bioinformatics. Nature Machine Intelligence  \textbf{2}(9),  500--508 (Sep
  2020)

\bibitem{Chen_2022_CVPR}
Chen, R.J., Chen, C., Li, Y., Chen, T.Y., Trister, A.D., Krishnan, R.G.,
  Mahmood, F.: Scaling vision transformers to gigapixel images via hierarchical
  self-supervised learning. In: Proceedings of the IEEE/CVF Conference on
  Computer Vision and Pattern Recognition (CVPR). pp. 16144--16155 (June 2022)

\bibitem{Chicco2020}
Chicco, D., Jurman, G.: The advantages of the matthews correlation coefficient
  (mcc) over f1 score and accuracy in binary classification evaluation. BMC
  Genomics  \textbf{21}(1), ~6 (Jan 2020)

\bibitem{Alfonso2021}
D'Alfonso, T.M., Ho, D.J., Hanna, M.G., Grabenstetter, A., Yarlagadda, D.V.K.,
  Geneslaw, L., Ntiamoah, P., Fuchs, T.J., Tan, L.K.: Multi-magnification-based
  machine learning as an ancillary tool for the pathologic assessment of shaved
  margins for breast carcinoma lumpectomy specimens. Modern Pathology
  \textbf{34}(8),  1487--1494 (Aug 2021)

\bibitem{https://doi.org/10.48550/arxiv.2012.03583}
Dehaene, O., Camara, A., Moindrot, O., de~Lavergne, A., Courtiol, P.:
  Self-supervision closes the gap between weak and strong supervision in
  histology (2020)

\bibitem{deng2009imagenet}
Deng, J., Dong, W., Socher, R., Li, L.J., Li, K., Fei-Fei, L.: Imagenet: A
  large-scale hierarchical image database. In: 2009 IEEE conference on computer
  vision and pattern recognition. pp. 248--255. Ieee (2009)

\bibitem{Eckardt2022}
Eckardt, J.N., Schmittmann, T., Riechert, S., Kramer, M., Sulaiman, A.S.,
  Sockel, K., Kroschinsky, F., Schetelig, J., Wagenf{\"u}hr, L., Schuler, U.,
  Platzbecker, U., Thiede, C., St{\"o}lzel, F., R{\"o}llig, C., Bornh{\"a}user,
  M., Wendt, K., Middeke, J.M.: Deep learning identifies acute promyelocytic
  leukemia in bone marrow smears. BMC Cancer  \textbf{22}(1), ~201 (Feb 2022)

\bibitem{ensemblelosslandscape}
Fort, S., Hu, H., Lakshminarayanan, B.: Deep ensembles: {A} loss landscape
  perspective. CoRR  \textbf{abs/1912.02757} (2019)

\bibitem{Ho_2021}
Ho, D.J., Yarlagadda, D.V., D'Alfonso, T.M., Hanna, M.G., Grabenstetter, A.,
  Ntiamoah, P., Brogi, E., Tan, L.K., Fuchs, T.J.: Deep multi-magnification
  networks for multi-class breast cancer image segmentation. Computerized
  Medical Imaging and Graphics  \textbf{88},  101866 (mar 2021)

\bibitem{Khened2021}
Khened, M., Kori, A., Rajkumar, H., Krishnamurthi, G., Srinivasan, B.: A
  generalized deep learning framework for whole-slide image segmentation and
  analysis. Scientific Reports  \textbf{11}(1),  11579 (Jun 2021)

\bibitem{adam}
Kingma, D.P., Ba, J.: Adam: A method for stochastic optimization (2014)

\bibitem{ensembleuncertainty}
Lakshminarayanan, B., Pritzel, A., Blundell, C.: Simple and scalable predictive
  uncertainty estimation using deep ensembles. In: {NIPS}. pp. 6402--6413
  (2017)

\bibitem{lee2019ensemble}
Lee, S., Amgad, M., Masoud, M., Subramanian, R., Gutman, D., Cooper, L.: An
  ensemble-based active learning for breast cancer classification. In: 2019
  IEEE International Conference on Bioinformatics and Biomedicine (BIBM). pp.
  2549--2553. IEEE (2019)

\bibitem{LeePCCB15}
Lee, S., Purushwalkam, S., Cogswell, M., Crandall, D.J., Batra, D.: Why {M}
  heads are better than one: Training a diverse ensemble of deep networks. CoRR
   \textbf{abs/1511.06314} (2015)

\bibitem{Li_2021_CVPR}
Li, B., Li, Y., Eliceiri, K.W.: Dual-stream multiple instance learning network
  for whole slide image classification with self-supervised contrastive
  learning. In: Proceedings of the IEEE/CVF Conference on Computer Vision and
  Pattern Recognition (CVPR). pp. 14318--14328 (June 2021)

\bibitem{LINDMAN201922}
Lindman, K., Rose, J.F., Lindvall, M., Lundstrom, C., Treanor, D.: Annotations,
  ontologies, and whole slide images – development of an annotated
  ontology-driven whole slide image library of normal and abnormal human
  tissue. Journal of Pathology Informatics  \textbf{10}(1), ~22 (2019)

\bibitem{Litjens20181399HS}
Litjens, G.J.S., B{\'a}ndi, P., Bejnordi, B.E., Geessink, O.G.F., Balkenhol,
  M.C.A., Bult, P., Halilovic, A., Hermsen, M., van~de Loo, R., Vogels, R.,
  Manson, Q.F., Stathonikos, N., Baidoshvili, A., van Diest, P., Wauters, C.A.,
  van Dijk, M., van~der Laak, J.: 1399 h\&e-stained sentinel lymph node
  sections of breast cancer patients: the camelyon dataset. GigaScience
  \textbf{7} (2018)

\bibitem{Liu2017}
Liu, Y., Gadepalli, K., Norouzi, M., Dahl, G.E., Kohlberger, T., Boyko, A.,
  Venugopalan, S., Timofeev, A., Nelson, P.Q., Corrado, G.S., Hipp, J.D., Peng,
  L., Stumpe, M.C.: Detecting cancer metastases on gigapixel pathology images
  (2017)

\bibitem{Lu2021}
Lu, M.Y., Williamson, D.F.K., Chen, T.Y., Chen, R.J., Barbieri, M., Mahmood,
  F.: Data-efficient and weakly supervised computational pathology on
  whole-slide images. Nature Biomedical Engineering  \textbf{5}(6),  555--570
  (Jun 2021)

\bibitem{mukherjee_emperor_2011}
Mukherjee, S.: The emperor of all maladies: a biography of cancer. Lions,
  London (2011), oCLC: 1031516848

\bibitem{Otsu1979ATS}
Otsu, N.: A threshold selection method from gray level histograms. IEEE
  Transactions on Systems, Man, and Cybernetics  \textbf{9},  62--66 (1979)

\bibitem{ritchie_causes_2019}
Ritchie, H., Spooner, F., Roser, M.: Causes of death (Dec 2019)

\bibitem{Schaumberg064279}
Schaumberg, A.J., Rubin, M.A., Fuchs, T.J.: H\&e-stained whole slide image deep
  learning predicts spop mutation state in prostate cancer. bioRxiv  (2018)

\bibitem{SCHUFFLER20219}
Schüffler, P.J., Yarlagadda, D.V.K., Vanderbilt, C., Fuchs, T.J.: Overcoming
  an annotation hurdle: Digitizing pen annotations from whole slide images.
  Journal of Pathology Informatics  \textbf{12}(1), ~9 (2021)

\bibitem{sung_global_2021}
Sung, H., Ferlay, J., Siegel, R.L., Laversanne, M., Soerjomataram, I., Jemal,
  A., Bray, F.: Global {Cancer} {Statistics} 2020: {GLOBOCAN} {Estimates} of
  {Incidence} and {Mortality} {Worldwide} for 36 {Cancers} in 185 {Countries}.
  CA: a cancer journal for clinicians  \textbf{71}(3),  209--249 (May 2021)

\bibitem{googlenet}
Szegedy, C., Liu, W., Jia, Y., Sermanet, P., Reed, S.E., Anguelov, D., Erhan,
  D., Vanhoucke, V., Rabinovich, A.: Going deeper with convolutions. In:
  {CVPR}. pp.~1--9. {IEEE} Computer Society (2015)

\bibitem{pmlr-v156-tourniaire21a}
Tourniaire, P., Ilie, M., Hofman, P., Ayache, N., Delingette, H.:
  Attention-based multiple instance learning with mixed supervision on the
  camelyon16 dataset. In: Proceedings of the MICCAI Workshop on Computational
  Pathology. Proceedings of Machine Learning Research, vol.~156, pp. 216--226.
  PMLR (27 Sep 2021)

\bibitem{https://doi.org/10.48550/arxiv.1606.05718}
Wang, D., Khosla, A., Gargeya, R., Irshad, H., Beck, A.H.: Deep learning for
  identifying metastatic breast cancer (2016)

\bibitem{labelcleaning}
Wang, Z., Popel, A.S., Sulam, J.: Label cleaning multiple instance learning:
  Refining coarse annotations on single whole-slide images. CoRR
  \textbf{abs/2109.10778} (2021)

\end{thebibliography}
% Added from bbl:

\appendix
\counterwithin{figure}{section}
\counterwithin{table}{section}

\section{Supplementary Material}

\begin{table*}[h!]
	\centering
 \resizebox{\textwidth}{!}{
	\begin{tabular}{c|ccc|cc}
		\hline
		Configuration    & Precision & Recall & Specificity & \textbf{\glsentryshort{auroc}} &\textbf{\glsentryshort{mcc}}\\
		\hline
		\multicolumn{6}{c}{\textbf{no noise}}\\
		\hline
		\glsentryshort{vit}-$5\times$ & $34.68 \pm 0.77$ & $71.97 \pm 1.3$ & $97.59 \pm 0.13$ & $92.79 \pm 0.29$ & $42.74 \pm 0.83$\\
		\glsentryshort{vit}-$10\times$ & $41.09 \pm 2.28$ & $79.12 \pm 0.99$ & $98.16 \pm 0.23$ & $95.67 \pm 0.17$ & $50.25 \pm 2.02$\\
		\glsentryshort{vit}-$20\times$ & $48.02 \pm 1.73$ & $80.23 \pm 0.81$ & $98.72 \pm 0.14$ & $96.3 \pm 0.1$ & $55.16 \pm 1.11$\\
		\glsentryshort{vit}-$40\times$ & $50.77 \pm 0.98$ & $68.48 \pm 0.97$ & $99.09 \pm 0.1$ & $93.83 \pm 0.26$ & $52.39 \pm 0.7$\\
		
		\hline
		\glsentryshort{ensemble} ($5\times^4$) & $36.66 \pm 0.31$ & $72.89 \pm 0.75$ & $97.89 \pm 0.04$ & $93.72 \pm 0.1$ & $45.1 \pm 0.5$\\
		\glsentryshort{ensemble} ($10\times^4$) & $43.64 \pm 0.66$ & $79.68 \pm 0.48$ & $98.47 \pm 0.06$ & $96.1 \pm 0.06$ & $52.59 \pm 0.5$\\
		\glsentryshort{ensemble} ($20\times^4$) & $51.26 \pm 0.52$ & $80.5 \pm 0.27$ & $98.97 \pm 0.03$ & $96.62 \pm 0.01$ & $57.71 \pm 0.31$\\
		\glsentryshort{ensemble} ($40\times^4$) & $53.6 \pm 0.26$ & $68.45 \pm 0.38$ & $99.25 \pm 0.02$ & $94.23 \pm 0.06$ & $54.39 \pm 0.18$\\
		\hline
		\glsentryshort{model} ($40, 20, 10$)  & $62.66 \pm 1.28$ & $79.01 \pm 0.58$ & $99.54 \pm 0.04$ & $97.76 \pm 0.06$ & $65.8 \pm 0.99$\\
		\glsentryshort{model} ($20, 10, 5$)& $54.82 \pm 1.57$ & $79.17 \pm 0.74$ & $99.13 \pm 0.05$ & $97.73 \pm 0.09$ & $60.99 \pm 1.18$\\
		\glsentryshort{model} ($40, 20, 10, 5$)& $64.15 \pm 1.22$ & $78.81 \pm 0.61$ & $99.44 \pm 0.01$ & $\bm{98.15 \pm 0.04}$ & $\bm{67.27 \pm 0.64}$\\
		\hline
		\multicolumn{6}{c}{\textbf{weak noise}}\\
		\hline
		\glsentryshort{vit}-$5\times$ & $31.77 \pm 0.89$ & $64.68 \pm 1.34$ & $97.42 \pm 0.19$ & $89.01 \pm 0.7$ & $35.83 \pm 0.81$\\
		\glsentryshort{vit}-$10\times$ & $34.09 \pm 0.5$ & $78.38 \pm 0.72$ & $96.87 \pm 0.15$ & $93.72 \pm 0.18$ & $42.6 \pm 0.5$\\
		\glsentryshort{vit}-$20\times$ & $40.57 \pm 1.15$ & $77.16 \pm 1.14$ & $98.05 \pm 0.2$ & $94.62 \pm 0.36$ & $47.36 \pm 0.95$\\
		\glsentryshort{vit}-$40\times$ & $42.56 \pm 0.89$ & $67.23 \pm 0.78$ & $98.08 \pm 0.13$ & $92.22 \pm 0.45$ & $44.71 \pm 0.57$\\
		\hline
		\glsentryshort{ensemble} ($5\times^4$) & $34.54 \pm 0.26$ & $65.05 \pm 0.55$ & $97.93 \pm 0.04$ & $90.07 \pm 0.13$ & $38.37 \pm 0.19$\\
		\glsentryshort{ensemble} ($10\times^4$) & $36.86 \pm 0.15$ & $78.51 \pm 0.28$ & $97.42 \pm 0.03$ & $94.34 \pm 0.02$ & $45.25 \pm 0.11$\\
		\glsentryshort{ensemble} ($20\times^4$) & $43.95 \pm 0.32$ & $77.36 \pm 0.3$ & $98.45 \pm 0.04$ & $95.23 \pm 0.07$ & $50.26 \pm 0.25$\\
		\glsentryshort{ensemble} ($40\times^4$) & $45.36 \pm 0.24$ & $67.09 \pm 0.13$ & $98.4 \pm 0.03$ & $92.9 \pm 0.1$ & $46.89 \pm 0.13$\\
		\hline
		
		\glsentryshort{model} ($40$, $20$, $10$)& 
		$55.04 \pm 0.6$ & $77.11 \pm 0.42$ & $99.23 \pm 0.05$ & $96.73 \pm 0.02$ & $58.76 \pm 0.43$\\
		\glsentryshort{model} ($20$, $10$, $5$) 
		& $50.03 \pm 1.03$ & $76.0 \pm 0.45$ & $98.97 \pm 0.07$ & $96.33 \pm 0.19$ & $54.95 \pm 0.93$\\
		\glsentryshort{model} ($40, 20, 10, 5$)& $58.47 \pm 1.02$ & $75.89 \pm 0.19$ & $99.35 \pm 0.02$ & $\bm{97.05 \pm 0.06}$ & $\bm{60.98 \pm 0.59}$\\
		
		\hline
		\multicolumn{6}{c}{\textbf{strong noise}}\\
		\hline
		\glsentryshort{vit}-$5\times$ & $29.04 \pm 0.75$ & $51.52 \pm 1.25$ & $97.55 \pm 0.18$ & $85.23 \pm 0.47$ & $28.98 \pm 0.58$\\
		\glsentryshort{vit}-$10\times$ & $31.27 \pm 0.8$ & $72.76 \pm 0.63$ & $96.25 \pm 0.27$ & $92.07 \pm 0.22$ & $37.87 \pm 0.75$\\
		\glsentryshort{vit}-$20\times$ & $34.65 \pm 1.0$ & $75.33 \pm 0.56$ & $96.72 \pm 0.2$ & $93.69 \pm 0.27$ & $41.29 \pm 0.82$\\
		\glsentryshort{vit}-$40\times$ & $36.48 \pm 0.68$ & $67.13 \pm 1.05$ & $96.55 \pm 0.14$ & $90.47 \pm 0.35$ & $38.99 \pm 0.27$\\
		\hline
		\glsentryshort{ensemble} ($5\times^4$) & $32.22 \pm 0.43$ & $51.94 \pm 0.91$ & $98.19 \pm 0.05$ & $86.8 \pm 0.13$ & $31.6 \pm 0.58$\\
		\glsentryshort{ensemble} ($10\times^4$) & $33.36 \pm 0.36$ & $72.81 \pm 0.18$ & $96.88 \pm 0.09$ & $92.86 \pm 0.05$ & $39.91 \pm 0.32$\\
		\glsentryshort{ensemble} ($20\times^4$) & $37.47 \pm 0.32$ & $75.71 \pm 0.16$ & $97.35 \pm 0.07$ & $94.49 \pm 0.06$ & $44.0 \pm 0.25$\\
		\glsentryshort{ensemble} ($40\times^4$) & $39.1 \pm 0.21$ & $67.15 \pm 0.23$ & $97.07 \pm 0.03$ & $91.32 \pm 0.09$ & $41.14 \pm 0.11$\\
		\hline
		
		\glsentryshort{model} ($40$, $20$, $10$)& 
		$49.38 \pm 0.82$ & $74.88 \pm 0.38$ & $98.75 \pm 0.06$ & $96.02 \pm 0.12$ & $52.96 \pm 0.38$\\
		\glsentryshort{model} ($20, 10, 5$) & 
		$47.68 \pm 1.38$ & $69.7 \pm 0.61$ & $98.88 \pm 0.07$ & $95.57 \pm 0.15$ & $49.84 \pm 0.88$\\
		\glsentryshort{model} ($40, 20, 10, 5$)  & $56.32 \pm 1.12$ & $71.28 \pm 0.42$ & $99.28 \pm 0.02$ & $\bm{96.43 \pm 0.13}$ & $\bm{56.63 \pm 0.87}$\\ 
		
		\hline
	\end{tabular}}
	\caption{Classification performance averaged over all \glspl{wsi} over five runs on the \gls{dataset} dataset for \glspl{model} and all baselines trained without noise, on weak and strong noise. $^4$ denotes single-magnification ensembles over 4 models acting on the same scale.}
	\label{tab:evalnoise}
\end{table*}

\begin{figure}[h!]
	\centering
	\subcaptionbox{\label{f2}}{%
		\includegraphics[width=0.51\textwidth]{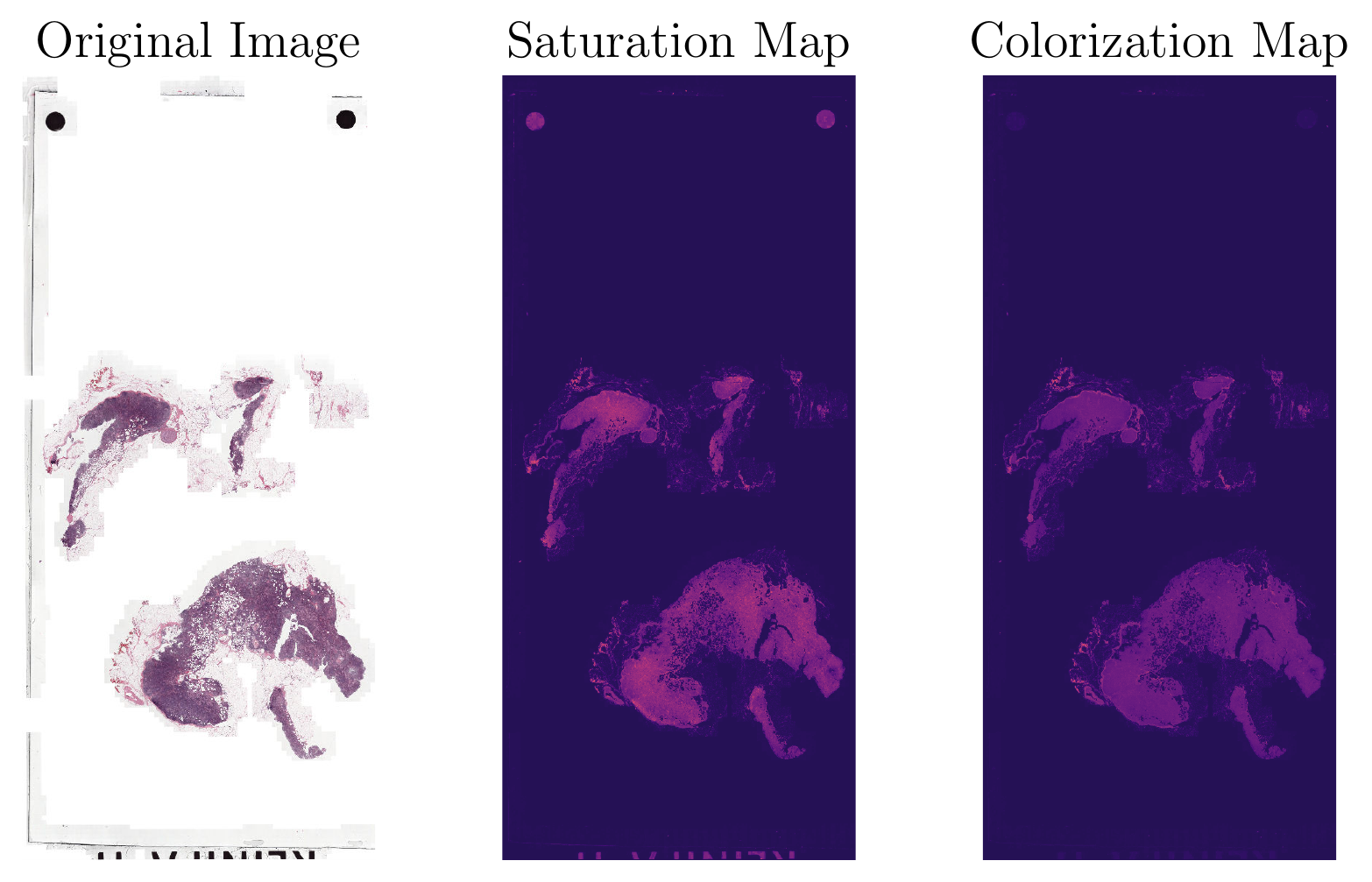}%
	}\quad
	\subcaptionbox{\label{s2}}{%
		\includegraphics[width=0.45\textwidth]{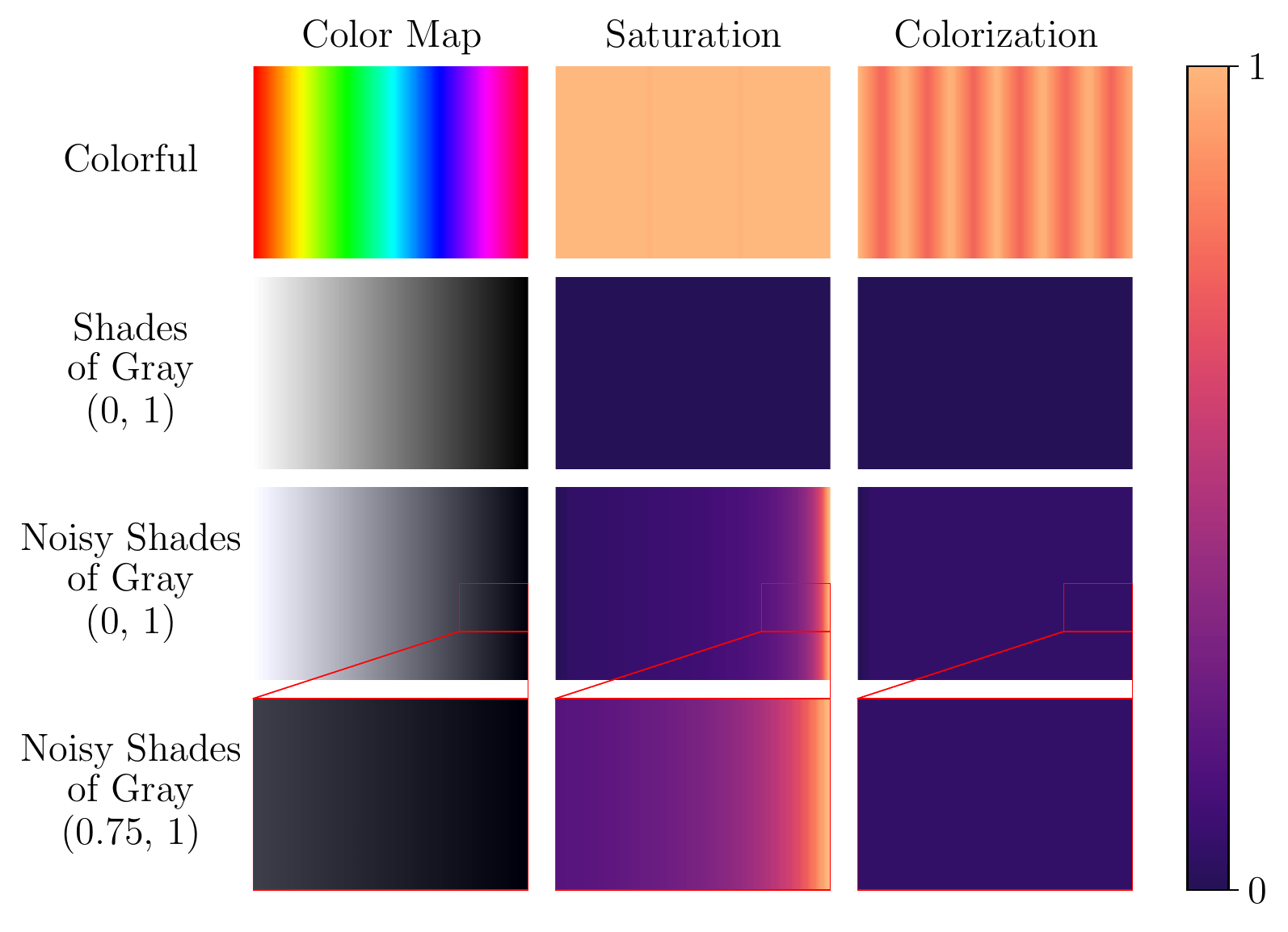}
	}
	
	\caption{(a) Illustration of a \gls{wsi} with its saturation, value and colorization maps. In contrast to saturation and value maps, colorization maps filter out both background and background noise, (b) The saturation values can be very high for noisy dark shades of grey, while the colorization values remain low for all shades of grey and high for all other colors.}
	\label{fig:segmentation}
\end{figure}
\begin{figure}[h!]
	\centering
	\includegraphics[width=1\textwidth]{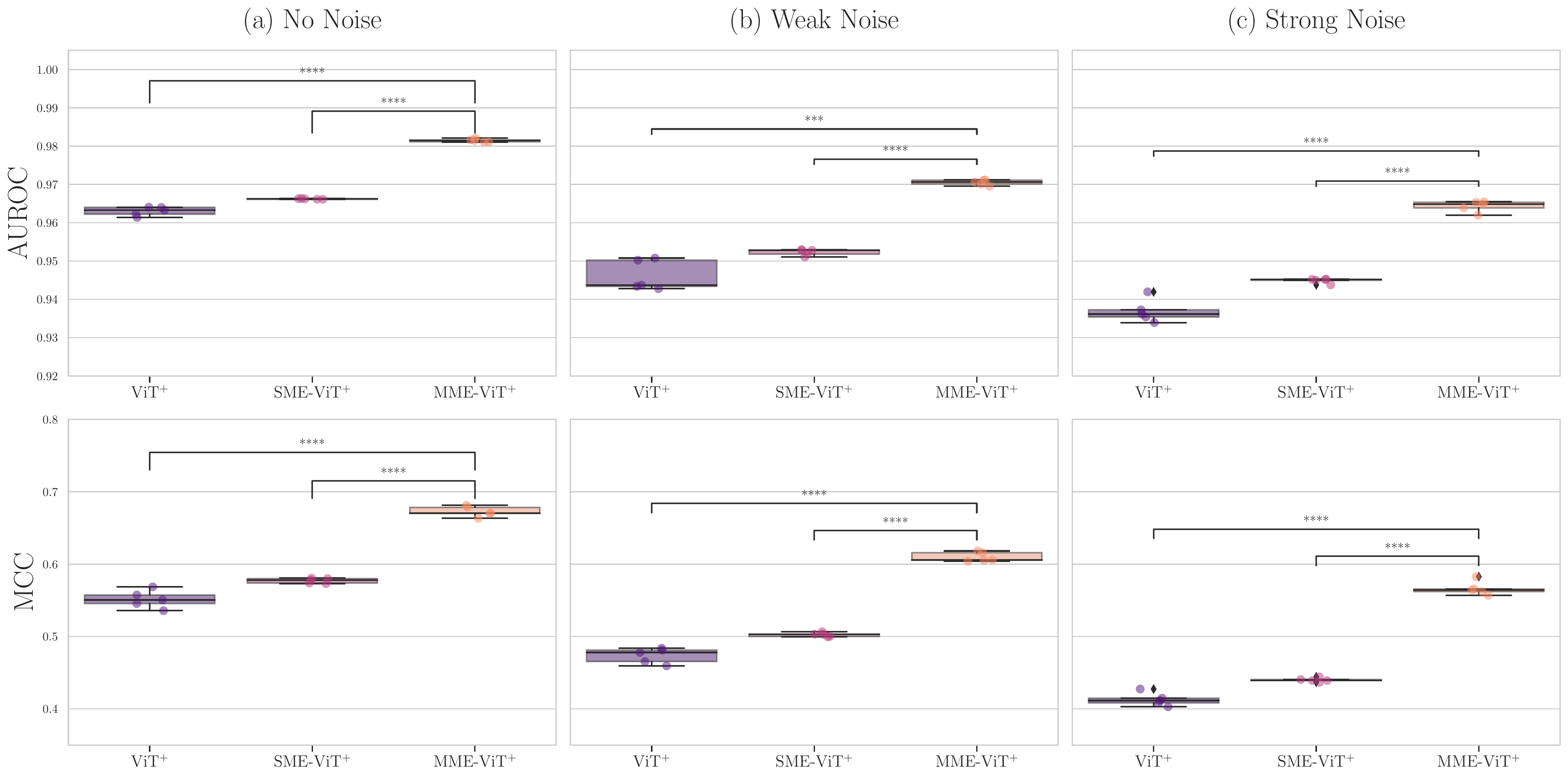}
	\caption{$^{**}$ denotes significance with $10^{-3} < p \leq 10^{-2}$, $^{***}$ with $10^{-4} < p \leq 10^{-3}$ and $^{****}$ with $p \leq 10^{-4}$ determined by Welch's t-test. We report the median and quartiles of the AUROC and MCC over five runs on the \gls{dataset} dataset for \gls{model} and the best single-magnification baselines trained (a) without artificial noise (b) with weak noise (c) with strong noise.}
	\label{fig:significance}
\end{figure}

\end{document}